# Measuring Semantic Similarity by Latent Relational Analysis

Peter D. Turney
Institute for Information Technology
National Research Council Canada
M-50 Montreal Road, Ottawa, Ontario, Canada, K1A 0R6
peter.turney@nrc-cnrc.gc.ca

## Abstract

This paper introduces Latent Relational Analysis (LRA), a method for measuring semantic similarity. LRA measures similarity in the semantic relations between two pairs of words. When two pairs have a high degree of relational similarity, they are analogous. For example, the pair cat:meow is analogous to the pair dog:bark. There is evidence from cognitive science that relational similarity is fundamental to many cognitive and linguistic tasks (e.g., analogical reasoning). In the Vector Space Model (VSM) approach to measuring relational similarity, the similarity between two pairs is calculated by the cosine of the angle between the vectors that represent the two pairs. The elements in the vectors are based on the frequencies of manually constructed patterns in a large corpus. LRA extends the VSM approach in three ways: (1) patterns are derived automatically from the corpus, (2) Singular Value Decomposition is used to smooth the frequency data, and (3) synonyms are used to reformulate word pairs. This paper describes the LRA algorithm and experimentally compares LRA to VSM on two tasks, answering college-level multiple-choice word analogy questions and classifying semantic relations in noun-modifier expressions. LRA achieves state-of-the-art results, reaching human-level performance on the analogy questions and significantly exceeding VSM performance on both tasks.

## 1 Introduction

This paper introduces Latent Relational Analysis (LRA), a method for measuring relational similarity. LRA has potential applications in many areas, including information extraction, word sense disambiguation, machine translation, and information retrieval.

*Relational similarity* is correspondence between relations, in contrast with *attributional similarity*, which is correspondence between attributes [Medin *et al.*, 1990]. When two words have a high degree of attributional similarity, we say they are *synonymous*. When two *pairs* of words have a high degree of relational similarity, we say they are *analogous*.

For example, the word pair mason:stone is analogous to the pair carpenter:wood; the relation between mason and stone is highly similar to the relation between carpenter and wood.

Past work on semantic similarity measures has mainly been concerned with attributional similarity. For instance, Latent Semantic Analysis (LSA) can measure the degree of similarity between two words, but not between two relations [Landauer and Dumais, 1997].

Recently the Vector Space Model (VSM) of information retrieval has been adapted to the task of measuring relational similarity, achieving a score of 47% on a collection of 374 college-level multiple-choice word analogy questions [Turney and Littman, 2005]. The VSM approach represents the relation between a pair of words by a vector of frequencies of predefined patterns in a large corpus.

LRA extends the VSM approach in three ways: (1) the patterns are derived automatically from the corpus (they are not predefined), (2) the Singular Value Decomposition (SVD) is used to smooth the frequency data (it is also used this way in LSA), and (3) automatically generated synonyms are used to explore reformulations of the word pairs. LRA achieves 56% on the 374 analogy questions, statistically equivalent to the average human score of 57%. On the related problem of classifying noun-modifier relations, LRA achieves similar gains over the VSM. For both problems, LRA's performance is state-of-the-art.

To motivate this research, Section 2 briefly outlines some possible applications for a measure of relational similarity. Related work with the VSM approach to relational similarity is described in Section 3. The LRA algorithm is presented in Section 4. LRA and VSM are experimentally evaluated by their performance on word analogy questions in Section 5 and on classifying semantic relations in noun-modifier expressions in Section 6. We discuss the interpretation of the results, limitations of LRA, and future work in Section 7. The paper concludes in Section 8.

## 2 Applications of Relational Similarity

Many problems in text processing would be solved (or at least greatly simplified) if we had a black box that could take as input two chunks of text and produce as output a measure of the degree of similarity in the meanings of the two chunks. We could use it for information retrieval, ques-tion answering, machine translation using parallel corpora, information extraction, word sense disambiguation, text summarization, measuring lexical cohesion, identifying sentiment and affect in text, and many other tasks in natural language processing. This is the vision that motivates research in paraphrasing [Barzilay and McKeown, 2001] and textual entailment [Dagan and Glickman, 2004], two topics that have lately attracted much interest.

In the absence of such a black box, current approaches to these problems typically use measures of attributional similarity. For example, the standard bag-of-words approach to information retrieval is based on attributional similarity [Salton and McGill, 1983]. Given a query, a search engine produces a ranked list of documents, where the rank of a document depends on the attributional similarity of the document to the query. The attributes are based on word frequencies; relations between words are ignored.

Although attributional similarity measures are very useful, we believe that they are limited and should be supplemented by relational similarity measures. Cognitive psychologists have also argued that human similarity judgements involve both attributional and relational similarity [Medin *et al.*, 1990].

Consider word sense disambiguation for example. In isolation, the word "plant" could refer to an industrial plant or a living organism. Suppose the word "plant" appears in some text near the word "food". A typical approach to disambiguating "plant" would compare the attributional similarity of "food" and "industrial plant" to the attributional similarity of "food" and "living organism" [Lesk, 1986; Banerjee and Pedersen, 2003]. In this case, the decision may not be clear, since industrial plants often produce food and living organisms often serve as food. It would be very helpful to know the relation between "food" and "plant" in this example. In the text "food *for* the plant", the relation between food and plant strongly suggests that the plant is a living organism, since industrial plants do not need food. In the text "food *at* the plant", the relation strongly suggests that the plant is an industrial plant, since living organisms are not usually considered as locations.

A measure of relational similarity could potentially improve the performance of any text processing application that currently uses a measure of attributional similarity. We believe relational similarity is the next step, after attributional similarity, towards the black box envisioned above.

## 3 Related Work

Let $R_1$ be the semantic relation between a pair of words, $A$ and $B$, and let $R_2$ be the semantic relation between another pair, $C$ and $D$. We wish to measure the relational similarity between $R_1$ and $R_2$. The relations $R_1$ and $R_2$ are not given to us; our task is to infer these hidden (latent) relations and then compare them.

In the VSM approach of Turney and Littman [2005], we create vectors, $r_1$ and $r_2$, that represent features of $R_1$ and $R_2$, and measure the similarity of $R_1$ and $R_2$ by the cosine of the angle $\theta$ between $r_1 = \langle r_{1,1},\ldots,r_{1,n} \rangle$ and $r_2 = \langle r_{2,1},\ldots r_{2,n} \rangle$:

$$\cos(\theta) = \frac{\sum_{i=1}^{n} r_{1,i} \cdot r_{2,i}}{\sqrt{\sum_{i=1}^{n} (r_{1,i})^2 \cdot \sum_{i=1}^{n} (r_{2,i})^2}} = \frac{r_1 \cdot r_2}{\|r_1\| \cdot \|r_2\|}.$$

We make a vector, $r$, to characterize the relationship between two words, $X$ and $Y$, by counting the frequencies of various short phrases containing $X$ and $Y$. Turney and Littman [2005] use a list of 64 joining terms, such as "of", "for", and "to", to form 128 phrases that contain $X$ and $Y$, such as *"X of Y", "Y of X", "X for Y", "Y for X", "X to Y",* and *"Y to X"*. These phrases are then used as queries for a search engine and the number of hits (matching documents) is recorded for each query. This process yields a vector of 128 numbers. If the number of hits for a query is $x$, then the corresponding element in the vector $r$ is $\log(x+1)$.

Turney and Littman [2005] evaluated the VSM approach by its performance on 374 college-level multiple-choice SAT analogy questions, achieving a score of 47%. A SAT analogy question consists of a target word pair, called the *stem*, and five *choice* word pairs. To answer an analogy question, vectors are created for the stem pair and each choice pair, and then cosines are calculated for the angles between the stem vector and each choice vector. The best guess is the choice pair with the highest cosine. We use the same set of analogy questions to evaluate LRA in Section 5.

The best previous performance on the SAT questions was achieved by combining thirteen separate modules [Turney *et al.*, 2003]. The performance of LRA significantly surpasses this combined system, but there is no real contest between these approaches, because we can simply add LRA to the combination, as a fourteenth module. Since the VSM module had the best performance of the thirteen modules [Turney *et al.*, 2003], the following experiments focus on comparing VSM and LRA.

The VSM was also evaluated by its performance as a distance measure in a supervised nearest neighbour classifier for noun-modifier semantic relations [Turney and Littman, 2005]. The problem is to classify a noun-modifier pair, such as "laser printer", according to the semantic relation between the head noun (printer) and the modifier (laser). The evaluation used 600 noun-modifier pairs that have been manually labeled with 30 classes of semantic relations [Nastase and Szpakowicz, 2003]. For example, "laser printer" is classified as *instrument*; the printer uses the laser as an instrument for printing. A testing pair is classified by searching for its single nearest neighbour in the labeled training data. The best guess is the label for the training pair with the highest cosine; that is, the training pair that is *most analogous* to the testing pair, according to VSM. LRA is evaluated with the same set of noun-modifier pairs in Section 6.

## 4 Latent Relational Analysis

LRA takes as input a set of word pairs and produces as output a measure of the relational similarity between any two of the input pairs. LRA relies on three resources, (1) a search engine with a very large corpus of text, (2) a broad-coverage thesaurus of synonyms, and (3) an efficient im-plementation of SVD. LRA does not use labeled data, structured data, or supervised learning.

LRA proceeds as follows:

**1. Find alternates:** For each word pair *A:B* in the input set, look in the thesaurus for the top num_sim words (in the following experiments, num_sim = 10) that are most similar to *A*. For each *A′* that is similar to *A*, make a new word pair *A′:B*. Likewise, look for the top num_sim words that are most similar to *B*, and for each *B′*, make a new word pair *A:B′*. *A:B* is called the *original* pair and each *A′:B* or *A:B′* is an *alternate* pair. The intent is for alternates to have almost the same semantic relations as the original.

**2. Filter alternates:** For each original pair *A:B*, filter the 2 × num_sim alternates as follows. For each alternate pair, send a query to the search engine, to find the frequency of phrases that begin with one member of the pair and end with the other. The phrases cannot have more than max_phrase words (we use max_phrase = 5). Sort the alternate pairs by the frequency of their phrases. Select the top num_filter most frequent alternates and discard the remainder (we use num_filter = 3, so 17 alternates are dropped). This step tends to eliminate alternates that have no clear semantic relation.

**3. Find phrases:** For each pair (originals and alternates), make a list of phrases in the corpus that contain the pair. Query the search engine for all phrases that begin with one member of the pair and end with the other, with a minimum of min_inter intervening words and a maximum of max_inter intervening words (we use min_inter = 1, max_inter = 3 = max_phrase − 2). We ignore suffixes when searching for phrases that match a given pair. These phrases reflect the semantic relations between the words in each pair.

**4. Find patterns:** For each phrase found in the previous step, build patterns from the intervening words. A pattern is constructed by replacing any or all or none of the intervening words with wild cards (one wild card can only replace one word). For each pattern, count the number of pairs (originals and alternates) with phrases that match the pattern (a wild card must match exactly one word). Keep the top num_patterns most frequent patterns and discard the rest (we use num_patterns = 4,000). Typically there will be millions of patterns, so it is not feasible to keep them all.

**5. Map pairs to rows:** In preparation for building a matrix $\mathbf{X}$, suitable for SVD, create a mapping of word pairs to row numbers. For each pair *A:B*, create a row for *A:B* and another row for *B:A*. This will make the matrix more symmetrical, reflecting our knowledge that the relational similarity between *A:B* and *C:D* should be the same as the relational similarity between *B:A* and *D:C*. (Mason is to stone as carpenter is to wood. Stone is to mason as wood is to carpenter.) The intent is to assist SVD by enforcing this symmetry in the matrix.

**6. Map patterns to columns:** Create a mapping of the top num_patterns patterns to column numbers. For each pattern *P*, create a column for "$word_1$ *P* $word_2$" and another column for "$word_2$ *P* $word_1$". Thus there will be 2 × num_patterns columns in $\mathbf{X}$.

**7. Generate a sparse matrix:** Generate a matrix $\mathbf{X}$ in sparse matrix format. The value for the cell in row *i* and column *j* is the frequency of the *j*-th pattern (see step 6) in phrases that contain the *i*-th word pair (see step 5).

**8. Calculate entropy:** Apply log and entropy transformations to the sparse matrix [Landauer and Dumais, 1997]. Each cell is replaced with its logarithm, multiplied by a weight based on the negative entropy of the corresponding column vector in the matrix. This gives more weight to patterns that vary substantially in frequency for each pair.

**9. Apply SVD:** After log and entropy transformations, apply SVD to $\mathbf{X}$. SVD decomposes $\mathbf{X}$ into a product of three matrices $\mathbf{U\Sigma V}^T$, where $\mathbf{U}$ and $\mathbf{V}$ are in column orthonormal form (i.e., the columns are orthogonal and have unit length) and $\mathbf{\Sigma}$ is a diagonal matrix of *singular values* (hence SVD) [Golub and Van Loan, 1996]. If $\mathbf{X}$ is of rank $r$, then $\mathbf{\Sigma}$ is also of rank $r$. Let $\mathbf{\Sigma}_k$, where $k < r$, be the diagonal matrix formed from the top $k$ singular values, and let $\mathbf{U}_k$ and $\mathbf{V}_k$ be the matrices produced by selecting the corresponding columns from $\mathbf{U}$ and $\mathbf{V}$. The matrix $\mathbf{U}_k\mathbf{\Sigma}_k\mathbf{V}_k^T$ is the matrix of rank $k$ that best approximates the original matrix $\mathbf{X}$, in the sense that it minimizes the approximation errors [Golub and Van Loan, 1996]. We may think of this matrix $\mathbf{U}_k\mathbf{\Sigma}_k\mathbf{V}_k^T$ as a "smoothed" or "compressed" version of the original matrix. SVD is used to reduce noise and compensate for sparseness.

**10. Projection:** Calculate $\mathbf{U}_k\mathbf{\Sigma}_k$ (we use $k = 300$, as recommended by Landauer and Dumais [1997]). This matrix has the same number of rows as $\mathbf{X}$, but only $k$ columns (instead of 2 × num_patterns columns; in our experiments, that is 300 columns instead of 8,000). We do not use $\mathbf{V}$, because we want to calculate the cosines between row vectors, and it can be proven that the cosine between any two row vectors in $\mathbf{U}_k\mathbf{\Sigma}_k$ is the same as the cosine between the corresponding two row vectors in $\mathbf{U}_k\mathbf{\Sigma}_k\mathbf{V}_k^T$.

**11. Evaluate alternates:** Let *A:B* and *C:D* be any two word pairs in the input set. From step 2, we have (num_filter + 1) versions of *A:B*, the original and num_filter alternates. Likewise, we have (num_filter + 1) versions of *C:D*. Therefore we have (num_filter + 1)² ways to compare a version of *A:B* with a version of *C:D*. Look for the row vectors in $\mathbf{U}_k\mathbf{\Sigma}_k$ that correspond to the versions of *A:B* and the versions of *C:D* and calculate the (num_filter + 1)² cosines (in our experiments, there are 16 cosines).

**12. Calculate relational similarity:** The relational similarity between *A:B* and *C:D* is the average of the cosines, among the (num_filter + 1)² cosines from step 11, that are greater than or equal to the cosine of the original pairs, *A:B* and *C:D*. The requirement that the cosine must be greater than or equal to the original cosine is a way of filtering out poor analogies, which may be introduced in step 1 and may have slipped through the filtering in step 2. Averaging the cosines, as opposed to taking their maximum, is intended to provide some resistance to noise.

In our experiments, the input set contains from 600 to 2,244 word pairs. Steps 11 and 12 can be repeated for each two input pairs that are to be compared.

In the following experiments, we use a local copy of the Waterloo MultiText System (WMTS) as the search engine, with a corpus of about $5 \times 10^{10}$ English words. The corpus was gathered by a web crawler from US academic web sites [Clarke *et al.*, 1998]. The WMTS is a distributed (multi-processor) search engine, designed primarily for passage retrieval (although document retrieval is possible, as a special case of passage retrieval). Our local copy runs on a 16-CPU Beowulf Cluster.

The WMTS is well suited to LRA, because it scales well to large corpora (one terabyte, in our case), it gives exact frequency counts (unlike most web search engines), it is designed for passage retrieval (rather than document retrieval), and it has a powerful query syntax.

As a source of synonyms, we use Lin's [1998] automatically generated thesaurus. Lin's thesaurus was generated by parsing a corpus of about $5 \times 10^7$ English words, consisting of text from the Wall Street Journal, San Jose Mercury, and AP Newswire [Lin, 1998]. The parser was used to extract pairs of words and their grammatical relations. Words were then clustered into synonym sets, based on the similarity of their grammatical relations. Two words were judged to be highly similar when they tended to have the same kinds of grammatical relations with the same sets of words.

Given a word and its part of speech, Lin's thesaurus provides a list of words, sorted in order of decreasing attributional similarity. This sorting is convenient for LRA, since it makes it possible to focus on words with higher attributional similarity and ignore the rest.

We use Rohde's SVDLIBC implementation of the Singular Value Decomposition, which is based on SVDPACKC [Berry, 1992].

## 5 Experiments with Word Analogy Questions

Table 1 shows one of the 374 SAT analogy questions, along with the relational similarities between the stem and each choice, as calculated by LRA. The choice with the highest relational similarity is also the correct answer for this question (quart is to volume as mile is to distance).

Table 1. Relation similarity measures for a sample SAT question.

| Stem:    |     | quart:volume   | Relational similarity |
|----------|-----|----------------|-----------------------|
| Choices: | (a) | day:night      | 0.373725              |
|          | (b) | mile:distance  | **0.677258**          |
|          | (c) | decade:century | 0.388504              |
|          | (d) | friction:heat  | 0.427860              |
|          | (e) | part:whole     | 0.370172              |

LRA correctly answered 210 of the 374 analogy questions and incorrectly answered 160 questions. Four questions were skipped, because the stem pair and its alternates did not appear together in any phrases in the corpus, so all choices had a relational similarity of zero. Since there are five choices for each question, we would expect to answer 20% of the questions correctly by random guessing. Therefore we score the performance by giving one point for each correct answer and 0.2 points for each skipped question. LRA attained a score of 56.4% on the 374 SAT questions.

The average performance of college-bound senior high school students on verbal SAT questions corresponds to a score of about 57% [Turney and Littman, 2005]. The difference between the average human score and the score of LRA is not statistically significant.

With 374 questions and 6 word pairs per question (one stem and five choices), there are 2,244 pairs in the input set. In step 2, introducing alternate pairs multiplies the number of pairs by four, resulting in 8,976 pairs. In step 5, for each pair *A:B*, we add *B:A*, yielding 17,952 pairs. However, some pairs are dropped because they correspond to zero vectors (they do not appear together in a window of five words in the WMTS corpus). Also, a few words do not appear in Lin's thesaurus, and some word pairs appear twice in the SAT questions (e.g., lion:cat). The sparse matrix (step 7) has 17,232 rows (word pairs) and 8,000 columns (patterns), with a density of 5.8% (percentage of nonzero values).

Table 2 compares LRA to VSM with the 374 analogy questions. VSM-AV refers to the VSM using AltaVista's database as a corpus. The VSM-AV results are taken from Turney and Littman [2005]. We estimate the AltaVista search index contained about $5 \times 10^{11}$ English words at the time the VSM-AV experiments took place. Turney and Littman [2005] gave an estimate of $1 \times 10^{11}$ English words, but we believe this estimate was slightly conservative. VSM-WMTS refers to the VSM using the WMTS, which contains about $5 \times 10^{10}$ English words. We generated the VSM-WMTS results by adapting the VSM to the WMTS.

Table 2. LRA versus VSM with 374 SAT analogy questions.

|           | VSM-AV | VSM-WMTS | LRA   |
|-----------|--------|----------|-------|
| Correct   | 176    | 144      | 210   |
| Incorrect | 193    | 196      | 160   |
| Skipped   | 5      | 34       | 4     |
| Total     | 374    | 374      | 374   |
| Score     | 47.3%  | 40.3%    | 56.4% |

All three pairwise differences in the three scores in Table 2 are statistically significant with 95% confidence, using the Fisher Exact Test. Using the same corpus as the VSM, LRA achieves a score of 56% whereas the VSM achieves a score of 40%, an absolute difference of 16% and a relative improvement of 40%. When VSM has a corpus ten times larger than LRA's corpus, LRA is still ahead, with an absolute difference of 9% and a relative improvement of 19%.

Comparing VSM-AV to VSM-WMTS, the smaller corpus has reduced the score of the VSM, but much of the drop is due to the larger number of questions that were skipped (34 for VSM-WMTS versus 5 for VSM-AV). With the smaller corpus, many more of the input word pairs simply do not appear together in short phrases in the corpus. LRA is able to answer as many questions as VSM-AV, although it uses the same corpus as VSM-WMTS, because Lin's [1998] thesaurus allows LRA to substitute synonyms for words that are not in the corpus.

VSM-AV required 17 days to process the 374 analogy questions [Turney and Littman, 2005], compared to 9 days for LRA. As a courtesy to AltaVista, Turney and Littman [2005] inserted a five second delay between each query. Since the WMTS is running locally, there is no need for delays. VSM-WMTS processed the questions in one day.

## 6 Experiments with Noun-Modifier Relations

This section describes experiments with 600 noun-modifier pairs, hand-labeled with 30 classes of semantic relations [Nastase and Szpakowicz, 2003]. We experiment with both a 30-class problem and a 5-class problem. The 30 classes of semantic relations include *cause* (e.g., in "flu virus", the head noun "virus" is the *cause* of the modifier "flu"), *location* (e.g., in "home town", the head noun "town" is the *location* of the modifier "home"), *part* (e.g., in "printer tray", the head noun "tray" is *part* of the modifier "printer"), and *topic* (e.g., in "weather report", the head noun "report" is about the *topic* "weather"). For a full list of classes, see Nastase and Szpakowicz [2003] or Turney and Littman [2005]. The 30 classes belong to 5 general groups of relations, *causal* relations, *temporal* relations, *spatial* relations, *participatory* relations (e.g., in "student protest", the "student" is the *agent* who performs the "protest"; *agent* is a *participatory* relation), and *qualitative* relations (e.g., in "oak tree", "oak" is a *type* of "tree"; *type* is a *qualitative* relation).

The following experiments use single nearest neighbour classification with leave-one-out cross-validation. For leave-one-out cross-validation, the testing set consists of a single noun-modifier pair and the training set consists of the 599 remaining noun-modifiers. The data set is split 600 times, so that each noun-modifier gets a turn as the testing word pair. The predicted class of the testing pair is the class of the single nearest neighbour in the training set. As the measure of nearness, we use LRA to calculate the relational similarity between the testing pair and the training pairs.

Following Turney and Littman [2005], we evaluate the performance by accuracy and also by the macroaveraged F measure [Lewis, 1991]. The F measure is the harmonic mean of precision and recall. Macroaveraging calculates the precision, recall, and F for each class separately, and then calculates the average across all classes.

There are 600 word pairs in the input set for LRA. In step 2, introducing alternate pairs multiplies the number of pairs by four, resulting in 2,400 pairs. In step 5, for each pair *A:B*, we add *B:A*, yielding 4,800 pairs. Some pairs are dropped because they correspond to zero vectors and a few words do not appear in Lin's thesaurus. The sparse matrix (step 7) has 4,748 rows and 8,000 columns, with a density of 8.4%.

Table 3 shows the performance of LRA and VSM on the 30-class problem. VSM-AV is VSM with the AltaVista corpus and VSM-WMTS is VSM with the WMTS corpus. The results for VSM-AV are taken from Turney and Littman [2005]. All three pairwise differences in the three F measures are statistically significant at the 95% level, according to the Paired T-Test. The accuracy of LRA is significantly higher than the accuracies of VSM-AV and VSM-WMTS, according to the Fisher Exact Test, but the difference between the two VSM accuracies is not significant. Using the same corpus as the VSM, LRA's accuracy is 15% higher in absolute terms and 61% higher in relative terms.

Table 4 compares the performance of LRA and VSM on the 5-class problem. The accuracy and F measure of LRA are significantly higher than the accuracies and F measures of VSM-AV and VSM-WMTS, but the differences between the two VSM accuracies and F measures are not significant. Using the same corpus as the VSM, LRA's accuracy is 14% higher in absolute terms and 32% higher in relative terms.

Table 3. Comparison of LRA and VSM on the 30-class problem.

|  | VSM-AV | VSM-WMTS | LRA |
| --- | --- | --- | --- |
| Correct | 167 | 148 | 239 |
| Incorrect | 433 | 452 | 361 |
| Total | 600 | 600 | 600 |
| Accuracy | 27.8% | 24.7% | 39.8% |
| Precision | 27.9% | 24.0% | 41.0% |
| Recall | 26.8% | 20.9% | 35.9% |
| F | 26.5% | 20.3% | 36.6% |

Table 4. Comparison of LRA and VSM on the 5-class problem.

|  | VSM-AV | VSM-WMTS | LRA |
| --- | --- | --- | --- |
| Correct | 274 | 264 | 348 |
| Incorrect | 326 | 336 | 252 |
| Total | 600 | 600 | 600 |
| Accuracy | 45.7% | 44.0% | 58.0% |
| Precision | 43.4% | 40.2% | 55.9% |
| Recall | 43.1% | 41.4% | 53.6% |
| F | 43.2% | 40.6% | 54.6% |

## 7 Discussion

The experimental results in Sections 5 and 6 demonstrate that LRA performs significantly better than the VSM, but it is also clear that there is room for improvement. The accuracy might not yet be adequate for practical applications, although past work has shown that it is possible to adjust the tradeoff of precision versus recall [Turney and Littman, 2005]. For some of the applications, such as information extraction, LRA might be suitable if it is adjusted for high precision, at the expense of low recall.

Another limitation is speed; it took almost nine days for LRA to answer 374 analogy questions. However, with progress in computer hardware, speed will gradually become less of a concern. Also, the software has not been optimized for speed; there are several places where the efficiency could be increased and many operations are parallelizable. It may also be possible to precompute much of the information for LRA, although this would require substantial changes to the algorithm.

The difference in performance between VSM-AV and VSM-WMTS shows that VSM is sensitive to the size of the corpus. Although LRA is able to surpass VSM-AV when the WMTS corpus is only about one tenth the size of the AV corpus, it seems likely that LRA would perform better with a larger corpus. The WMTS corpus requires one terabyte of hard disk space, but progress in hardware will likely make ten or even one hundred terabytes affordable in the relatively near future.

For noun-modifier classification, more labeled data should yield performance improvements. With 600 noun-modifier pairs and 30 classes, the average class has only 20 examples. We expect that the accuracy would improve sub-stantially with five or ten times more examples, but it is time consuming and expensive to acquire hand-labeled data.

Another issue with noun-modifier classification is the choice of classification scheme for the semantic relations. The 30 classes of Nastase and Szpakowicz [2003] might not be the best scheme. Other researchers have proposed different schemes [Rosario and Hearst, 2001]. It seems likely that some schemes are easier for machine learning than others.

# 8 Conclusion

This paper has introduced a new method for calculating relational similarity, Latent Relational Analysis. The experiments demonstrate that LRA performs better than the VSM approach, when evaluated with SAT word analogy questions and with the task of classifying noun-modifier expressions. The VSM approach represents the relation between a pair of words with a vector, in which the elements are based on the frequencies of 64 hand-built patterns in a large corpus. LRA extends this approach in three ways: (1) the patterns are generated dynamically from the corpus, (2) SVD is used to smooth the data, and (3) a thesaurus is used to explore reformulations of the word pairs.

Just as attributional similarity measures have proven to have many practical uses, we expect that relational similarity measures will soon become widely used. Relational similarity plays a fundamental role in the mind and therefore relational similarity measures could be crucial for artificial intelligence [Medin *et al.*, 1990]. LRA may be a step towards the black box that we imagined in Section 2, with many potential applications in text processing.

In future work, we plan to investigate some potential applications for LRA. It is possible that the error rate of LRA is still too high for practical applications, but the fact that LRA matches average human performance on SAT analogy questions is encouraging.

## Acknowledgments

For data and software used in this research, my thanks go to Michael Littman, Vivi Nastase, Stan Szpakowicz, Egidio Terra, Charlie Clarke, Dekang Lin, Doug Rohde, and Michael Berry.